\pdfoutput=1

\documentclass[11pt]{article}

\usepackage{acl}
\usepackage{times}
\usepackage{latexsym}
\usepackage[T1]{fontenc}
\usepackage[utf8]{inputenc}
\usepackage{microtype}

\usepackage{amsmath,amssymb}
\usepackage{adjustbox}
\usepackage{booktabs}
\usepackage{comment}
\usepackage[shortlabels]{enumitem} 
    \setitemize{noitemsep,topsep=0em} 
    \setenumerate{noitemsep,itemindent=13pt}
\usepackage{footnote}
\usepackage{flushend}
\usepackage{graphicx}
\usepackage{multirow}
\usepackage{todonotes}
\usepackage{xspace}

\newcommand*{\flores}{\textsc{Flores}\xspace}
\definecolor{rscolor}{RGB}{158, 25, 70}

\newcommand{\Coolname}{Pre-trained multilingual sequence-to-sequence}

\newcommand{\coolabbrev}{PMSS}

\newcommand{\lang}[1]{\textsc{#1}}

\usepackage[noabbrev,capitalize,nameinlink]{cleveref}
\crefname{section}{\S}{\S\S}
\Crefname{section}{\S}{\S\S}    
\crefformat{section}{#2\S#1#3}  
\Crefformat{section}{#2\S#1#3}
\crefrangeformat{section}{\S\S#3#1#4--#5#2#6}
\Crefrangeformat{section}{\S\S#3#1#4--#5#2#6}
\crefformat{section}{#2\S#1#3}  
\crefrangeformat{section}{\S\S#3#1#4--#5#2#6}

\def\Snospace~{\S{}}

\title{ Pre-Trained  Multilingual Sequence-to-Sequence Models:\\
A Hope for Low-Resource Language Translation?}

\author{
     \mbox{\bf En-Shiun Annie Lee\rlap{\textnormal{,}}}\textsuperscript{*} 
    \mbox{Sarubi Thillainathan\rlap{\textnormal{,}}}\textsuperscript{$\dagger$}
    \mbox{Shravan Nayak\rlap{\textnormal{,}}}\textsuperscript{$\ddagger$}
    \mbox{Surangika Ranathunga\rlap{\textnormal{,}}}\textsuperscript{$\dagger$} \\ 
    \mbox{\bf David Ifeoluwa Adelani\rlap{\textnormal{,}}}\textsuperscript{$\mathsection\text{,}\mathparagraph$}
    \mbox{\bf Ruisi Su\rlap{\textnormal{,}}}\textsuperscript{$\|$} \and
    \mbox{\bf Arya D. McCarthy}\textsuperscript{\#} \\
    \textsuperscript{*}University of Toronto,
     \textsuperscript{$\dagger$}University of Moratuwa, \textsuperscript{$\ddagger$}IIT(BHU) Varanasi,\\  \textsuperscript{$\mathsection$}Masakhane NLP, \textsuperscript{$\mathparagraph$}Saarland University, \textsuperscript{$\|$}Sway AI,  \textsuperscript{\#}Johns Hopkins University 
     \\
     \texttt{annie.lee@cs.toronto.edu}\\
}

\begin{document}
\maketitle
\begin{abstract}
    What can pre-trained multilingual sequence-to-sequence models like mBART contribute to translating low-resource languages? 
    We conduct a thorough empirical experiment in 10 languages  to ascertain this, considering five factors: (1) the amount of fine-tuning data, (2) the noise in the fine-tuning data, (3) the amount of pre-training data in the model, (4) the impact of domain mismatch, and (5) language typology. 
    In addition to yielding several heuristics, the experiments form a framework for evaluating the data sensitivities of machine translation systems.
    While mBART is robust to domain differences, its translations for unseen and typologically distant languages remain below 3.0 BLEU.
    In answer to our title's question, mBART is not a low-resource panacea; we therefore encourage shifting the emphasis from new models to new data\footnote{Code is available at \url{https://github.com/LRLNMT/LRLNMT}}.\looseness=-1 
\end{abstract}

\section{Introduction}

\Coolname{} (\coolabbrev{}) models, such as mBART \citep{tang2020multilingual} and mT5 \citep{xue2021mt5}, are pre-trained on large general data, 
then fine-tuned to deliver impressive 
results for natural language inference, question answering, and text simplification \citep{xtreme}. Their performance on machine translation shows promise for translating low-resource languages \citep{liu-etal-2021-continual,adelani-etal-2021-effect,thillainathan2021fine}, which remains an open challenge \citep{bitext,koehn-knowles-2017-six,mager-etal-2021-findings,ranathunga2021neural}.

When can mBART and mT5 succeed in translating a low-resource language? 
Despite their promise, the specific conditions for their practical application are not yet clear.
Understanding their sensitivities is crucial to guide data acquisition efforts and apply \coolabbrev{} models to new languages.

We introduce a framework for assessing data-dependency of performance of machine translation systems.
We then apply it in a
large-scale study of mBART's viability for low-resource machine translation on 10 typologically and geographically varied languages. Eight languages are low-resource, and four are unseen by mBART during pre-training. Through our results, we gauge the importance of five dimensions of the training data:
\begin{enumerate}[topsep=0.4em]
    \item Amount of fine-tuning data
    \item Noise in fine-tuning data
    \item Amount of pre-training data
    \item Domain mismatch
    \item Language typology
\end{enumerate}
The closest work to ours \citep{liu-etal-2021-continual} considers only the first two.

For the seen languages, mBART reaches acceptable 
performance with either 10k high-quality, in-domain sentence pairs or 100k noisy ones. However, mBART's BLEU score for unseen languages is often below 3.0---far below usability. 
For these unseen, low-resource languages, the fact that even mBART---which has already seen billions of sentences---cannot succeed in virtually any of our conditions speaks to the need for appropriate in-domain data.
Therefore, the analytical framework in our experimental design can help to target new data acquisition efforts.

\section{Models and Data}
mBART and mT5 are \coolabbrev{} models that rely on the encoder--decoder Transformer architecture \citep{vaswani2017attention} trained on Common Crawl--derived data with variants of a monolingual autoencoding objective: they must recreate the input text that they are provided. 
Neither is trained with an explicit objective encouraging similar tokens or sentences to have similar representations.

After model weights have been learned, the models can be fine-tuned on parallel text for translation. 
The ideal fine-tuning scenario would be vast, clean data matching the language and domain of interest.
Because this scenario is unlikely for low-resource languages, we test the relaxation of these assumptions for \coolabbrev{} models.

In a preliminary experiment comparing mBART and mT5
, mBART performed better than mT5 on 11 of the 18 translation directions, especially the \lang{en}\(\to\)xx directions (\cref{tab:mbartvsmt5}), corroborating \citet{liu-etal-2021-continual}. 
Because mBART performed better both in number of translation directions and average BLEU, we focus hereafter on it.

\begin{table}[t]
\centering
\tiny
\begin{tabular}{@{} l l r r r r r @{}} \toprule \multicolumn{1}{l}{} & \multicolumn{1}{l}{} & \multicolumn{1}{l}{} & \multicolumn{2}{c}{\lang{en}→xx}     & \multicolumn{2}{c}{xx→\lang{en}}     \\
\cmidrule(lr){4-5} \cmidrule(l){6-7}
Language             & Training data              & Size                 & mBART         & mT5           & mBART         & mT5           \\ \midrule

\lang{af}                   & JW300                & 1,104k                 & 30.9          & \textbf{32.9} & 43.9          & \textbf{46.9} \\
\lang{xh}                   & JW300                & 866k                 & \textbf{9.1}  & 8.4           & 22.8          & \textbf{23.2} \\
\lang{yo}                   & JW300                & 472k               & \textbf{3.9}  & 2.6           & 7.9           & \textbf{8.1}  \\
\lang{ga}                   & EUBookShop           & 133k                  & \textbf{15.1} & 7.6           & 15.7          & \textbf{16.7} \\ 
\lang{fr}                   & DGT-TM           & 100k                  & \textbf{18.8} &       19.8    &    19.3      &  \textbf{20.3} \\
\lang{\lang{si}}                   & Gov't                 & 56k                 & \textbf{5.4}  & 2.3           & \textbf{9.6}  & 8.4           \\
\lang{ta}                   & Gov't                 & 56k                  & \textbf{3.5}  & 2.4           & \textbf{10.7} & 10.1          \\
\lang{hi}                   & PMIndia           & 50k                  & \textbf{14.1} & 10.5          & \textbf{19.5}          & 16.4 \\ 
\lang{kn}                   & PMIndia           & 25k                  & \textbf{4.1} & 2.9          & 4.2          & \textbf{10.7} \\

\addlinespace
Average                   &                 &                & \textbf{11.7}  & 9.9           & 17.1           & \textbf{17.9}  \\

\bottomrule
\end{tabular}

\caption{Preliminary results for mBART and mT5 (base version) in six languages. We test on \textsc{Flores} in all cases. The best score for each direction is in bold.}
\label{tab:mbartvsmt5}
\vspace{-1mm}
\end{table}

\subsection{Languages}
To assess mBART's translation ability, we selected a set of high- and low-resource languages with high typological and geographical diversity (\cref{tab:lang-details}).  Five of the ten languages do not use the Latin script, so that we can evaluate mBART's generalization to non-Latin scripts \citep[see][]{pires-etal-2019-multilingual}. Eight are considered low-resource languages by \citet{joshi-etal-2020-state}, while two high-resource languages (\lang{fr} and \lang{hi}) give a skyline of performance.\footnote{\citet{joshi-etal-2020-state}'s taxonomy is out-of-date. Because \lang{si} is used to train mBART, it must be at least class 3. We believe that, according to \citet{joshi-etal-2020-state}'s definition, no language in our study is below class 2.} 
Four are unseen during mBART's pre-training. 
Together, these languages let us probe the effects of pre-training data size and language typology on translation.

\begin{table}[t]
\centering
\small
\begin{adjustbox}{max width=\linewidth}
\begin{tabular}{ l@{\quad}lllrr  @{}}
\toprule
& & &     & Joshi & mBART  \\
\multicolumn{2}{l}{Language} & Family    & Script    & class  & tokens  \\\midrule
\lang{fr} & French       & Romance (IE)     & Latin     & 5 & 9780M       \\
\lang{hi} & Hindi       & Indo-Aryan (IE)  & Devanagari & 4 & 1715M       \\
\lang{ta} & Tamil       & Dravidian   & Tamil   & 3   & 595M       \\
\lang{si} & Sinhala       & Indo-Aryan (IE)  & Sinhala  & 1  & 243M       \\
\lang{af} & Afrikaans       & Germanic (IE)    & Latin    & 3  & 242M       \\
\lang{xh} & Xhosa       & Niger--Congo & Latin    & 2  & 13M       \\
\lang{ga} & Irish       & Celtic (IE)       & Latin    & 2  & --       \\ 
\lang{yo} & Yor\`ub\'a       & Niger--Congo & Latin    & 2  & --        \\
\lang{as} & Assamese       & Indo-Aryan (IE)  & Bengali--Assamese       & 1   & --        \\
\lang{kn} & Kannada       & Dravidian       & Kannada  & 1  & --        \\

\bottomrule
\end{tabular}
\end{adjustbox}

\caption{The 10 languages in our study.}
\label{tab:lang-details}
\vspace{-1mm}
\end{table}

\subsection{Corpora}
Selecting suitable parallel corpora enables us to probe the remaining three factors: amount of fine-tuning data, noise in the fine-tuning data, and domain mismatch.

For each of our 10 languages, we use three training corpora: data from Common Crawl, the Bible, and one other domain-specific dataset (\cref{tab:data-details}; complete details  in \cref{sec:data}).
Common Crawl is large and open-domain, while the others are smaller curated translations. We use \flores{} (which is also open-domain) and the two domain-specific corpora for testing.  
Comparing on these lets us assess the impact of domain mismatch.

To evaluate consistently across differently sized corpora, we sampled fixed-size training sets from each corpus. For the Common Crawl data, we used two sizes: 25k and 100k sentence pairs. For the Bible, we used a 1k-sentence-pair sample. Finally, for each language's other domain-specific dataset, depending on the amount of parallel text available, we used up to four sizes (1k, 10k, 50k, 100k).

The Common Crawl datasets are large open-domain parallel corpora, but their construction by automatic alignment invites substantial noise. This problem is especially severe for low-resource languages \citep{qualityAtAGlance}. Noisy data often harm translation models \citep{khayrallah-koehn-2018-impact}, but it is possible to use them effectively \citep{mccarthy-etal-2020-addressing}.
This raises the question of whether mBART can do so.
Among our experiments, we can see whether and when a smaller, clean parallel corpus would be preferable.

\begin{table}[t]
\centering
\begin{adjustbox}{max width=\linewidth}
\footnotesize
\begin{tabular}{@{} l @{\qquad\qquad} l @{\qquad\qquad} l @{}} \toprule
Dataset               & Domain & Languages \\ \midrule
\textsc{Flores}-101                & Open       & all except \lang{si}          \\
\textsc{Flores}v1                & Open       & \lang{si}          \\
\addlinespace
CCAligned             & Open       & all except \lang{ga}               \\
CCMatrix              & Open       &  \lang{ga}                      \\
\addlinespace
JHU Bibles                 & Religious       &  all                   \\
\addlinespace
JW300                 & Religious+magazines      & \lang{af}, \lang{yo}, \lang{xh}            \\
Government            & Administrative       & \lang{\lang{si}}, \lang{ta}               \\
PMIndia               & News       & \lang{as}, \lang{kn}, \lang{hi}            \\
DGT-TM                & Legal       & \lang{fr}, \lang{ga}             \\\bottomrule
\end{tabular}
\end{adjustbox}
\caption{Parallel corpora used in our study.}
\label{tab:data-details}
\vspace{-1mm}
\end{table}

\definecolor{mygray}{gray}{0.8}
\newcommand{\gray}[1]{\textcolor{mygray}{#1}}
\newlength\datgap
\setlength{\datgap}{2em}
\newcommand{\push}{\hspace{1.5em}}
\begin{table*}[ht!]
\centering
\resizebox{\textwidth}{!}{%
\begin{tabular}{@{} lr @{\hspace{\datgap}} rrr @{\hspace{\datgap}} rrr @{\hspace{\datgap}} rrr @{\hspace{\datgap}} rrr @{\hspace{\datgap}} rrr @{\hspace{\datgap}} rrr @{}}
\toprule
& & \multicolumn{9}{c}{\lang{\lang{en}}→xx} & \multicolumn{9}{c}{xx→\lang{\lang{en}}} \\
\cmidrule(r{\datgap}){3-11} \cmidrule(){12-20}
  & & \multicolumn{3}{c}{\lang{af}} & \multicolumn{3}{c}{\lang{xh}} & \multicolumn{3}{c}{\lang{yo}} & \multicolumn{3}{c}{\lang{af}} & \multicolumn{3}{c}{\lang{xh}} & \multicolumn{3}{c}{\lang{yo}} \\
\cmidrule(r{\datgap}){3-5} \cmidrule(r{\datgap}){6-8} \cmidrule(r{\datgap}){9-11} \cmidrule(r{\datgap}){12-14} \cmidrule(r{\datgap}){15-17} \cmidrule(){18-20}
Training & Size & \textsc{Flores} & Bible & JW300 & \textsc{Flores} & Bible & JW300 & \textsc{Flores}  & Bible & JW300 & \textsc{Flores} & Bible & JW300 & \textsc{Flores} & Bible & JW300 & \textsc{Flores}  & Bible & JW300 \\
\midrule
\textit{Transformer} \\
\push Bible & 1k & \gray{0.1} & 1.3 & \gray{0.7} & \gray{0.0} & \gray{0.0} & \gray{0.0} & \gray{0.0} & 1.4 & \gray{0.0} & \gray{0.1} & 1.7 & \gray{0.8} & \gray{0.0} & \gray{0.9} & \gray{0.2} & \gray{0.0} & 2.4 & \gray{0.0} \\
\push JW300 & 100k & \textbf{19.2} & \textbf{13.8}  & \textbf{44.2} & 1.8 & \gray{0.7} & \textbf{31.8} & 1.2 & \gray{0.6}   & \textbf{18.7}  & \textbf{22.5}   & \textbf{15.1}  & \textbf{42.4} & 6.6 & 4.9   & \textbf{37.5}  & 2.4     & 1.0   & \textbf{17.7}  \\
\push Common Crawl & 100k & \textbf{23.6} & 7.0 & \textbf{17.4} & 2.5 & \gray{0.6} & 2.3 & 1.2 & 1.6 & 1.4 & \textbf{28.3} & \textbf{10.3} & \textbf{22.3}  & 7.7 & 2.9 & \textbf{10.2}  & 2.1 & 3.3 & 4.1 \\
\addlinespace

\textit{mBART50} \\
\push Bible & 1k & \gray{0.1}    & \gray{0.1}   & \gray{0.1}   & \gray{0.6}    & \gray{0.2}   & 3.5   & \gray{0.6}     & 3.6   & 3.6   & \textbf{20.5}   & \textbf{13.4}  & \textbf{23.5}  & 2.8    & 3.3   & 3.1   & \gray{0.2}     & \gray{0.4}   & \gray{0.2}   \\
\addlinespace
\push JW300 & 1k  & \textbf{18.9}   & \textbf{11.1}  & \textbf{32.4}  & 1.6    & \gray{0.1}   & \textbf{11.0}  & 1.0     & \gray{0.0}   & 6.7   & \textbf{28.8}   & \textbf{12.6}  & \textbf{32.5}  & \gray{0.1}    & \gray{0.1}   & \gray{0.1}   & \gray{0.0}     & \gray{0.0}   & \gray{0.0}   \\
& 10k  & \textbf{26.5}   & \textbf{14.1}  & \textbf{42.7}  & 4.1    & 1.8   & \textbf{22.1}  & 2.0     & \gray{0.2}   & 7.8   & \textbf{32.4}   & \textbf{16.0}  & \textbf{39.0}  & \textbf{11.4}   & 4.8   & \textbf{29.1}  & 6.2     & 1.0   & \textbf{15.4}  \\
& 50k & \textbf{30.1}   & \textbf{15.8}  & \textbf{48.0}  & 6.0    & 4.0   & \textbf{30.8}  & 3.8     & \gray{0.7}   & \textbf{20.1}  & \textbf{40.9}   & \textbf{17.5}  & \textbf{41.7}  & \textbf{16.2}   & 9.2   & \textbf{41.3}  & 7.8     & 1.3   & \textbf{19.8}  \\
& 100k & \textbf{30.1}   & \textbf{16.2}  & \textbf{49.7}  & 7.4    & 4.3   & \textbf{34.9}  & 3.9     & \gray{0.9}   & \textbf{23.6}  & \textbf{42.0}   & \textbf{17.9}  & \textbf{43.7}  & \textbf{19.9}   & \textbf{11.5}  & \textbf{45.7}  & 7.9     & 1.5   & \textbf{22.0}  \\ 
\addlinespace
\push Common Crawl & 25k & \textbf{28.0} & \textbf{13.4}  & \textbf{31.4}  & 4.8    & \gray{0.5}   & \textbf{10.1}  & 2.6     & 1.7   & 3.8   & \textbf{36.0}   & \textbf{15.0}  & \textbf{35.0}  & \textbf{11.3}   & 3.0   & \textbf{18.6}  & 3.5     & 3.2   & 5.2 \\

& 100k & \textbf{33.9} & \textbf{15.5}  & \textbf{34.4}  & 7.9    & 2.1   & \textbf{16.8}  & 2.8     & 4.5   & 5.9   & \textbf{44.8}   & \textbf{16.9}  & \textbf{40.2}  & \textbf{19.7}   & 9.0   & \textbf{27.8}  & 5.0     & 7.5   & 6.7  \\
\bottomrule

\toprule
& & \multicolumn{9}{c}{\lang{\lang{en}}→xx} & \multicolumn{9}{c}{xx→\lang{\lang{en}}} \\
\cmidrule(r{\datgap}){3-11} \cmidrule(){12-20}
 &  & \multicolumn{3}{c}{\lang{hi}} & \multicolumn{3}{c}{\lang{kn}} & \multicolumn{3}{c}{\lang{as}} & \multicolumn{3}{c}{\lang{hi}} & \multicolumn{3}{c}{\lang{kn}} & \multicolumn{3}{c}{\lang{as}} \\
\cmidrule(r{\datgap}){3-5} \cmidrule(r{\datgap}){6-8} \cmidrule(r{\datgap}){9-11} \cmidrule(r{\datgap}){12-14} \cmidrule(r{\datgap}){15-17} \cmidrule(){18-20}
Training & Size & \textsc{Flores} & Bible & PMI & \textsc{Flores} & Bible & PMI & \textsc{Flores}  & Bible & PMI & \textsc{Flores} & Bible & PMI & \textsc{Flores} & Bible & PMI & \textsc{Flores}  & Bible & PMI \\
\midrule
\textit{Transformer} \\
\push Bible & 1k & \gray{0.0} & \gray{0.2}   & \gray{0.0}     & \gray{0.0}     & \gray{0.0}   & \gray{0.0}     & \gray{0.0}     & \gray{0.0}    & \gray{0.0}    & \gray{0.0}     & \gray{0.0}   & \gray{0.0}       & \gray{0.0}     & \gray{0.9}    & \gray{0.0}      & \gray{0.0}     & \gray{0.3}    & \gray{0.0}   \\
\push PMI & 50k & 7.7 & 1.3   & \textbf{22.9}  & \gray{0.0}     & \gray{0.0}   & 4.9   & \gray{0.0}     & \gray{0.0}    & 1.3  & 7.7     & 2.4   & \textbf{26.2}    & 6.6     & \gray{0.6}    & 9.7    & \gray{0.0}     & \gray{0.0}    & 3.4  \\
\push Common Crawl & 100k & 8.7 & 2.3   & 7.3   & \gray{0.2} & \gray{0.0} & \gray{0.0}   & \gray{0.0}     & \gray{0.0}    & \gray{0.0}  & 6.6     & 3.0   & 4.7     & \gray{0.1}     & \gray{0.0}    & \gray{0.1}    & \gray{0.0}     & \gray{0.1}    & \gray{0.1}  \\
\addlinespace
\textit{mBART50} \\
\push Bible & 1k & 3.7     & 7.0   & 4.3   & \gray{0.0}     & \gray{0.1}   & \gray{0.0}   & \gray{0.1}     & \gray{0.9}    & -    & 7.1     & 9.3   & 7.2     & \gray{0.1}     & \gray{0.3}    & \gray{0.0}    & 1.4     & 4.6    & -     \\
\addlinespace
\push PMI & 1k & 7.0	    & 2.3	& \textbf{14.5}  & \gray{0.0}     & \gray{0.0}   & \gray{0.1}   & \gray{0.0} 	& \gray{0.0}	 & 2.1  & 7.4	  & 4.1	  & \textbf{11.8}    & \gray{0.3}	  & \gray{0.1}	   & 1.7    & \gray{0.0}	 & \gray{0.0}	   & \gray{0.2}    \\
& 10k & \textbf{11.5}    & 2.5   & \textbf{24.2}  & 1.8     & \gray{0.1}   & \textbf{10.7}  & -       & -      & -    & \textbf{16.8}	  & 7.1	  & \textbf{30.6}    & \gray{0.9}	  & \gray{0.2}    & 5.2    & -       & -      & -  \\
& 50k & \textbf{14.1}	& 3.4	& \textbf{28.8}  & -       & -     & -     & -       & -      & -    & \textbf{19.5}	  & 8.2   & \textbf{37.6}    & -       & -      & -      & -       & -      & -     \\
\addlinespace
\push Common Crawl & 25k & \textbf{14.2}    & 5.5   & \textbf{12.0}  & \gray{0.4}     & \gray{0.0}   & \gray{0.1}   & 1.4     & \gray{0.3}    & 1.4  & \textbf{17.6}    & \textbf{10.2}  & \textbf{14.0}    & \gray{0.2}     & \gray{0.0}    & \gray{0.1}    & 1.6     & \gray{0.8}    & 1.6  \\
& 100k & \textbf{20.9}    & 6.2   & \textbf{17.0}  & 1.2     & \gray{0.0}   & \gray{0.7}   & -       & -      & -    & \textbf{22.4}    & \textbf{11.2}  & \textbf{17.1}    & \gray{0.4}     & \gray{0.0}    & \gray{0.5}    & -       & -      & -   \\
\bottomrule

\toprule
& & \multicolumn{9}{c}{\lang{\lang{en}}→xx} & \multicolumn{9}{c}{xx→\lang{\lang{en}}} \\
\cmidrule(r{\datgap}){3-11} \cmidrule(){12-20}
  & & \multicolumn{3}{c}{\lang{si}} & \multicolumn{3}{c}{\lang{ta}} & \multicolumn{3}{c}{\lang{ga}} & \multicolumn{3}{c}{\lang{si}} & \multicolumn{3}{c}{\lang{ta}} & \multicolumn{3}{c}{\lang{ga}} \\
\cmidrule(r{\datgap}){3-5} \cmidrule(r{\datgap}){6-8} \cmidrule(r{\datgap}){9-11} \cmidrule(r{\datgap}){12-14} \cmidrule(r{\datgap}){15-17} \cmidrule(){18-20}
Training & Size & \textsc{Flores} & Bible & Gov't & \textsc{Flores} & Bible & Gov't & \textsc{Flores}  & Bible & DGT & \textsc{Flores} & Bible & Gov't & \textsc{Flores} & Bible & Gov't & \textsc{Flores}  & Bible & DGT \\
\midrule
\textit{Transformer} \\

\push Bible  & 1k & \gray{0.0} & \gray{0.0}   & \gray{0.0}       & \gray{0.0}     & \gray{0.0}   & \gray{0.0}     & \gray{0.0}     & \gray{0.1}    & \gray{0.0}        & \gray{0.0}     & 1.1   & \gray{0.1}     & \gray{0.0}       & \gray{0.7}   & \gray{0.0}   & \gray{0.0}     & 1.0    & \gray{0.0}   \\
\push Gov't/DGT & 50k/100k  & 1.3     & \gray{0.0}   & \textbf{20.6}      & \gray{0.5}     & \gray{0.0}   & \textbf{13.7}    & 3.3     & \gray{0.0}    & 3.2        & 2.7     & \gray{0.4}   & \textbf{23.9}    & 2.7     & \gray{0.7}   & \textbf{23.9}    & 3.2     & \gray{0.0}    & 3.0  \\
\push Common Crawl & 100k  & 2.1     & \gray{0.0}   & 5.6       & 1.8     & \gray{0.0}   & 1.8     & \gray{0.0}     & \gray{0.0}    & \gray{0.0}         & 4.7     & 1.9   & 7.9     & 5.2     & 3.4   & 4.9    & \gray{0.1}     & \gray{0.0}    & \gray{0.0}  \\
\addlinespace
\textit{mBART50} \\
\push Bible & 1k & \gray{0.2}     & 3.6   & 1.2       & \gray{0.7}     & 1.1   & 1.1     & \gray{0.9}     & 1.3    & \gray{0.1}       & 4.8     & 9.0   & 4.5     & 5.3     & 7.8   & 4.4      & \gray{0.0}     & \gray{0.0}    & \gray{0.0}  \\
\addlinespace
\push Gov't/DGT & 1k & 1.4     & \gray{0.1}   & \textbf{11.2}      & 1.1     & \gray{0.1}   & 6.6     & \gray{0.8}     & \gray{0.0}    & 1.5         & 6.5     & 2.5   & \textbf{14.8}    & 6.1     & 2.1   & \textbf{12.6}   & \gray{0.3}     & \gray{0.1}    & \gray{0.8}   \\
& 10k & 4.2     & \gray{0.2}   & \textbf{26.4}      & 2.3     & \gray{0.2}   & \textbf{17.4}    & 4.7     & \gray{0.1}    & 4.1        & 8.4     & 3.3   & \textbf{30.7}    & 7.7     & 2.6   & \textbf{23.8}    & 5.8     & \gray{0.2}    & 4.7  \\
& 50k & 5.1     & \gray{0.2}   & \textbf{35.4}      & 3.7     & \gray{0.2}   & \textbf{23.4}    & \textbf{12.2}    & \gray{0.3}    & 4.2        & 9.2     & 3.5   & \textbf{38.8}    & \textbf{10.4}    & 3.3   & \textbf{37.3}    & \textbf{12.3}    & \gray{0.4}    & 5.1  \\ 
& 100k & -       & -     & -         & -       & -     & -       & 8.9     & \gray{0.2}    & 4.3           & -       & -   & -    & -    & -  & -               & 9.5     & \gray{0.2}    & 4.9  \\
\addlinespace
\push Common Crawl & 25k & 4.4     & \gray{0.5}   & 9.6       & 4.7     & \gray{0.9}   & 4.6     & \gray{0.0}     & \gray{0.0}    & \gray{0.0}        & 9.6     & 5.2   & \textbf{13.5}    & 7.2     & 6.5   & 5.6     & \gray{0.1}     & \gray{0.1}    & \gray{0.0}    \\
& 100k & 6.6     & \gray{0.5}   & \textbf{16.9}      & 7.6     & \gray{0.8}   & 8.6     & \gray{0.0}     & \gray{0.0}    & \gray{0.0}        & \textbf{13.8}    & 8.5   & \textbf{20.5}    & \textbf{17.3}    & 9.6   & \textbf{16.8}    & \gray{0.0}     & \gray{0.0}    & \gray{0.0}  \\
\bottomrule

\end{tabular}%
}
\vspace{-1mm}
\caption{Experimental results, reported in SacreBLEU~\cite{post-2018-call}. Values <1.0 grey; values >10.0 bold.}
\label{tab:result-lk-eu}
\end{table*}

\section{Experimental Setting}

We fine-tune mBART models on each of the training corpora and sizes listed above, and we evaluate their performance using the development and test sets from the domain-specific corpora and \flores{}.

We additionally train a standard Transformer baseline \citep{vaswani2017attention} to compare pre-training versus training from scratch.

We score translations with SacreBLEU \citep{post-2018-call}. Details of training and evaluation are given in \cref{app:experiments}.

\section{Results and Analysis}
\begin{table}[t]
\centering
\begin{adjustbox}{max width=\linewidth}
\begin{tabular}{@{} lr @{\hspace{3em}} rrr @{\hspace{3em}} rrr @{}} \toprule
\multicolumn{1}{l}{} & \multicolumn{1}{l}{} & \multicolumn{3}{c}{\lang{en}→\lang{fr}} & \multicolumn{3}{c}{\lang{fr}→\lang{en}} \\
\cmidrule(r{3em}){3-5} \cmidrule(){6-8}
Training            & Size                 & \flores   & Bible  & DGT   & \flores   & Bible  & DGT   \\ \midrule
\textit{Transformer}\\
\push Bible                & 1k                   & \gray{0.0}      & 2.4    & \gray{0.0}   & \gray{0.0}      & 1.6    & \gray{0.0}   \\
\push DGT                  & 100k                 & 5.7      & 1.4    & \textbf{22.8}  & 6.1      & 2.4    & \textbf{26.6}  \\ 
\push Common Crawl                   & 100k                 & 9.0      & 6.5    & 5.6   & \textbf{10.7}     & 6.8    & 7.3   \\
\addlinespace
\textit{mBART50} \\
\push Bible                & 1k                   & \textbf{13.2}     & \textbf{15.5}   & \textbf{10.9}  & \gray{0.0}      & \gray{0.0}    & \gray{0.0}   \\
\addlinespace
\push DGT                  & 1k                   & \textbf{15.1}     & 5.7    & \textbf{20.2}  & \textbf{19.9}     & \textbf{11.9}   & \textbf{27.8}  \\
& 10k                  & \textbf{15.5}     & 4.4    & \textbf{25.4}  & \textbf{17.7}     & 7.8    & \textbf{29.7}  \\
& 50k                  & \textbf{17.8}     & 5.1    & \textbf{31.2}  & \textbf{18.3}     & 8.5    & \textbf{35.3}  \\

& 100k                 & \textbf{18.8}     & 5.0    & \textbf{34.6}  & \textbf{19.3}     & 7.6    & \textbf{36.6} \\
\addlinespace
\push Common Crawl                   & 25k                  & \textbf{24.0}     & \textbf{14.9}   & \textbf{15.6}  & \textbf{26.0}     & \textbf{18.0}   & \textbf{19.4}  \\
& 100k                 & \textbf{29.4}     & \textbf{16.3}   & \textbf{19.6}  & \textbf{29.1}     & \textbf{18.9}   & \textbf{22.6}  \\
\bottomrule
\end{tabular}%
\end{adjustbox}
\vspace{-1mm}
\caption{Experimental results for French, reported in SacreBLEU. Values <1.0 grey; values >10.0 bold.}
\label{tab:result-fr}
\end{table}

The results of our empirical study are given in \cref{tab:result-lk-eu}, with \lang{fr} given in \cref{tab:result-fr}. By contrasting specific groups of rows, we probe our five factors.

\subsection{Amount of fine-tuning data} \label{sec:fine-tuning}

To assess this dimension, we compare the Transformer and mBART models trained on varying sizes of the same corpus with their corresponding open-domain and domain-specific evaluation sets.

In the open-domain case (training on Common Crawl), for languages seen during pre-training, mBART fine-tuned with 25k sentence pairs outperforms the Transformer trained with 100k parallel sentences; this pattern holds for 18 of the 20 language directions. This indicates that pre-trained mBART is at least four times as data-efficient. Although it also outperforms the Transformer on unseen languages in terms of BLEU, the scores are often below 3.0---a far cry from even the BLEU score needed for gisting.

\begin{figure}[t]
\includegraphics[width=\linewidth]{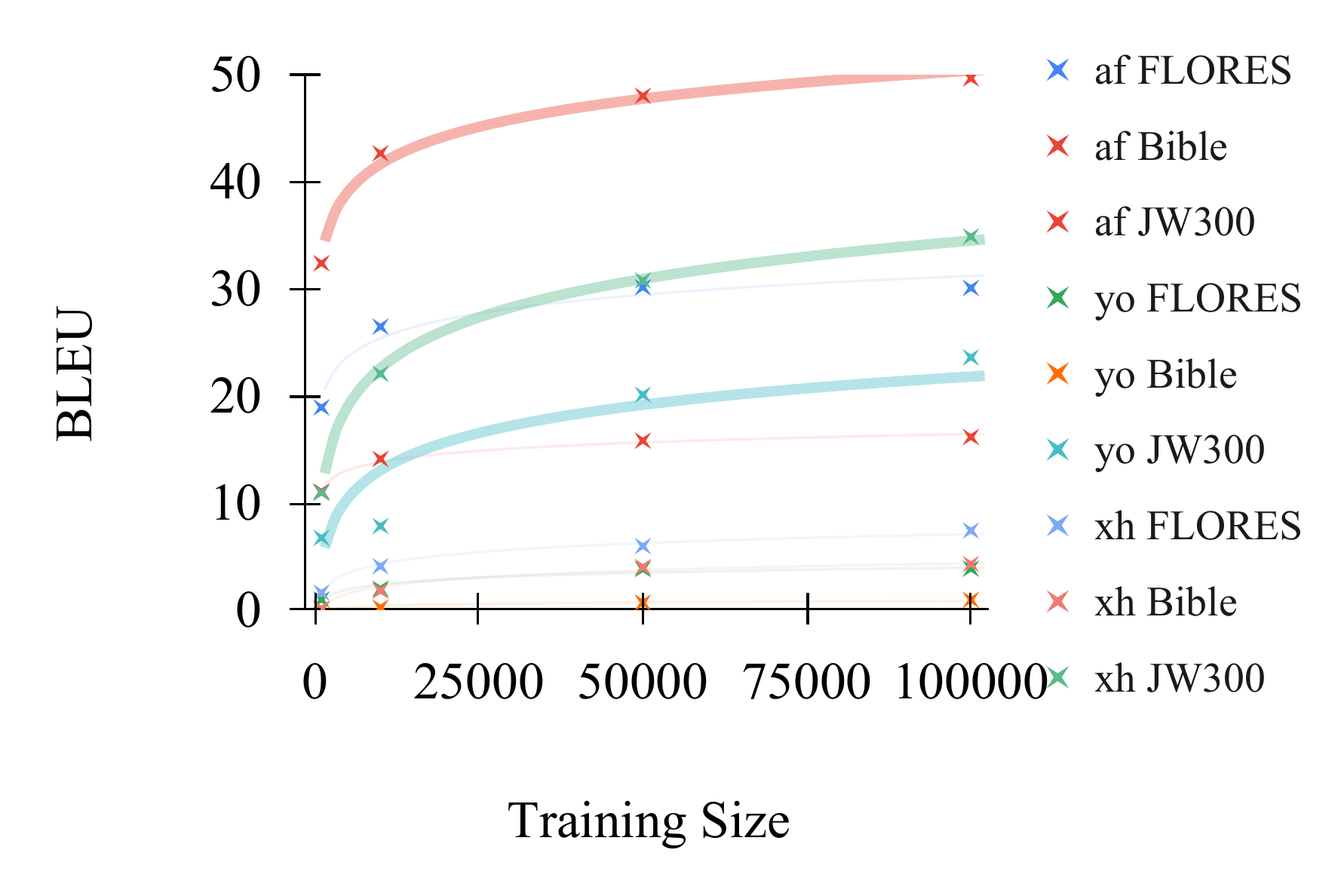}
\caption{Impact of fine-tuning dataset size on mBART performance translating into English on JW300.}
\label{fig:fine-tuning_size}
\vspace{-2mm}
\end{figure}
On the other hand, we observe a similar trend when training with domain-specific datasets (JW300, Gov't, and DGT).
For the government-domain dataset, mBART trained with 10k sentences of \lang{si} or \lang{ta} achieves a higher BLEU than the Transformer trained with 50k sentences (+3.4 to +6.8); this suggests at least a fivefold data efficiency. The exception is \lang{si}\(\to\)\lang{en}, where the difference in scores is 0.1 BLEU.
For JW300, mBART trained with 10k parallel sentences outperforms the Transformer trained with 100k for some translation tasks tenfold.
Further, mBART trained with 50k sentences outperforms the Transformer model for all languages by a large margin\footnote{The only exceptions are \textit{\lang{\lang{af}-\lang{en}}} and
\textit{\lang{\lang{en}}-\lang{xh}} in-domain testing, with less than or equal to 1.0 BLEU point difference.}.
Of note, \lang{yo} begins to perform well in-domain on JW300 with tens of thousands of sentences.

When do we reach diminishing returns on fine-tuning size? \Cref{fig:fine-tuning_size} shows how fine-tuning size affects translation of JW300 into \lang{en} from \lang{af}, \lang{xh}, and \lang{yo}. Although training with more data improves BLEU, the gain saturates as the dataset size reaches approximately 50k sentence pairs. \citeauthor{liu2020multilingual_mbart25} attribute this to the limit of the model's capacity: that the pre-trained weights are ``washed out'' \citeyearpar{liu2020multilingual_mbart25} when fine-tuning with more parallel data.

\subsection{Noise in fine-tuning data}
At what point is a small-but-clean corpus more useful than an automatically mined one like from Common Crawl? Comparing mBART trained on Common Crawl versus domain-specific data, we see that for several languages both in and not in mBART, 10k high-quality in-domain sentences leads to better performance than 100k sentences from Common Crawl.

\subsection{Amount of pre-training data}
\begin{figure}[t]
\includegraphics[width=\linewidth]{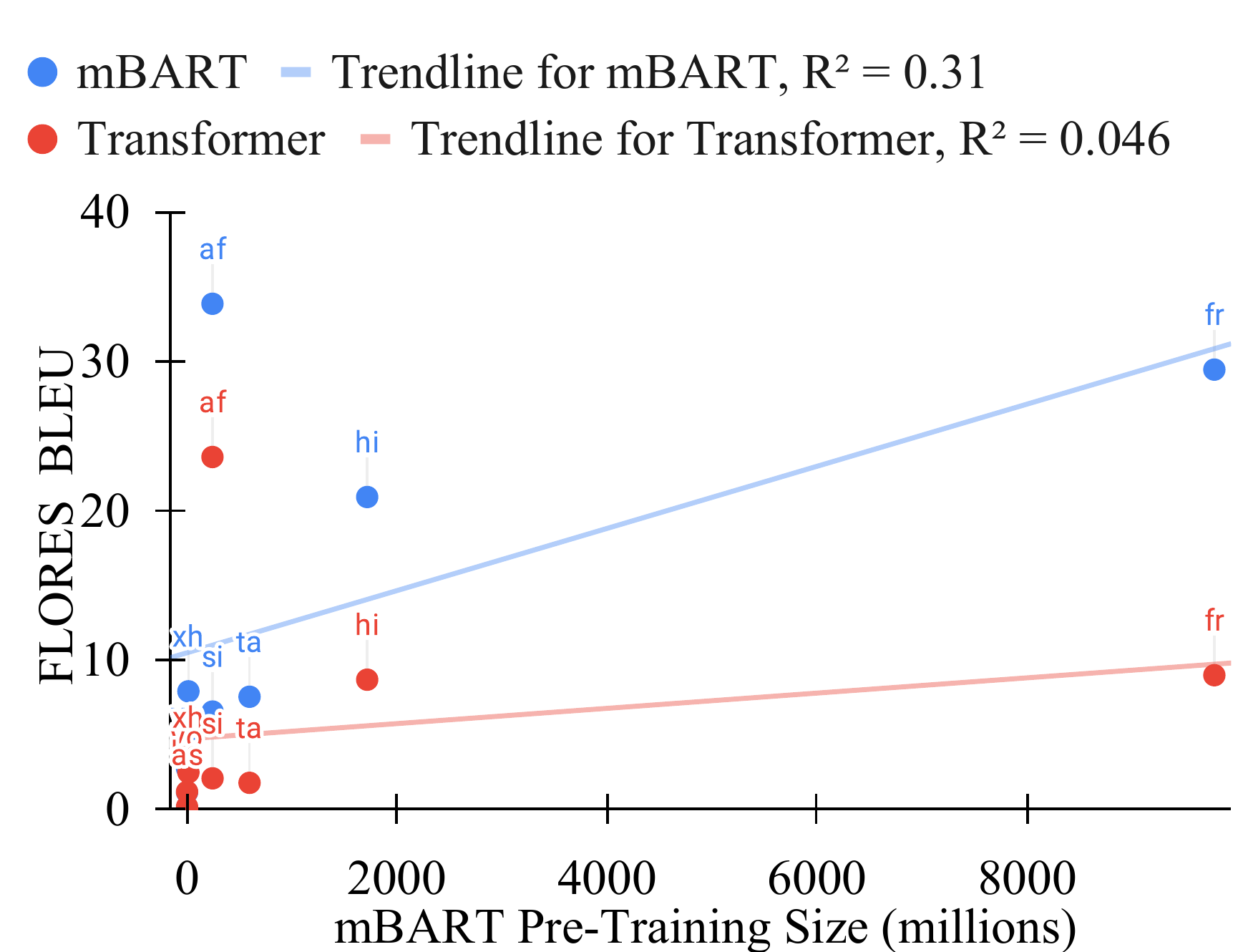}
\caption{Effect of pre-training open-domain dataset size, using 100k Common Crawl sentence pairs for fine-tuning, translating from English}
\label{fig:pre-training_size}
\vspace{-2mm}
\end{figure}

The improvement of mBART over the Transformer is more prominent for languages with more pre-training data.
The correlation between BLEU and number of pre-training sentences is \(R^2 = 0.31\) for open-domain (\cref{fig:pre-training_size}), and the effect in the domain-specific case is similar.
This shows that mBART effectively leverages the pre-training data. 
Taken with the results of \cref{sec:fine-tuning},
the contrasting behavior between seen and unseen languages belies a ``rich-get-richer'' phenomenon.

\subsection{Domain mismatch}

This section compares the performance of models when trained and tested on matching versus mismatched domains. 


Unsurprisingly, taking a training set from the same domain as the test set consistently yields higher BLEU than a mismatched training set. This pattern repeats across domains and directions. 

Of greater interest is that Common Crawl--trained models often do better on domain-specific test sets than open-domain test sets. For languages with JW300 or Gov't, testing BLEU on these was higher than on the open-domain \flores{} data.

Further, for \lang{si} and \lang{ta}, mBART
trained on 10k sentences achieved higher BLEU
 than the Transformer trained
on 100k data, suggesting the pre-training gain
was able to compensate the lack of in-domain
data.  This may indicate that mBART is valuable for domain-specific translation with low amounts of high-quality data.

Results for \lang{fr} on DGT and the Bible and \lang{hi} on PMI show that mBART can excel with even 1k parallel sentences for languages with sufficient pre-training.
If data from a different domain is available in sufficient quantities, an acceptable translation can be expected, as evident from the Gov't 50k and JW300 100k settings. 
Noticeably, issues related to domain difference and fine-tuning dataset size are less pronounced for \textit{\lang{fr}} (see results for 1k Bible data and 1k DGT). This reiterates the impact of language coverage in the mBART model.

\subsection{Language typology}
\begin{figure}
    \centering
    \includegraphics[width=1\linewidth]{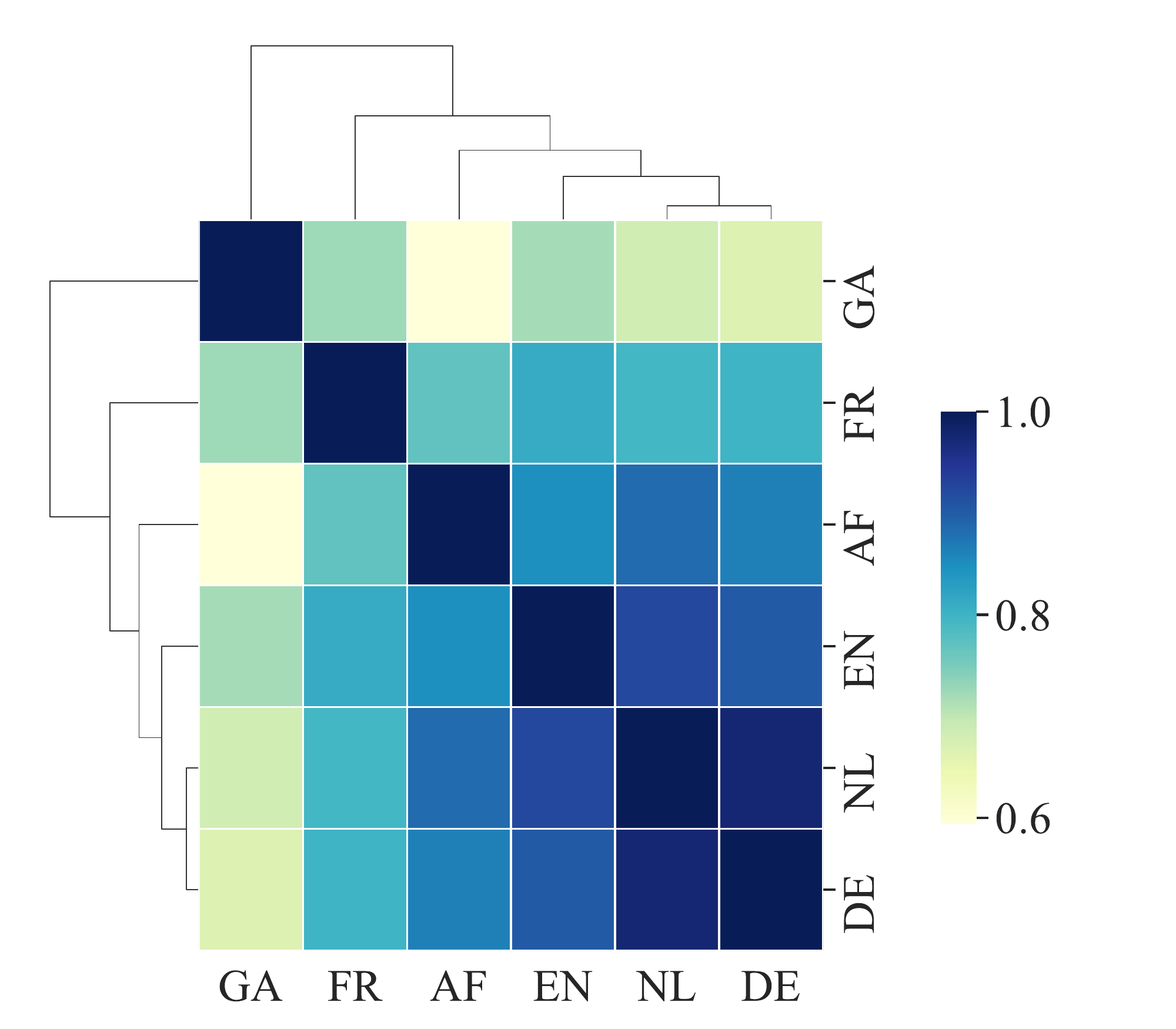}
    \caption{Cosine similarities of syntactic features}
    \label{fig:syn_sim}
\end{figure}

This analysis relates properties of the languages to their performance.

Foremost, \lang{af} regularly achieves the highest BLEU among low-resource languages used to pre-train mBART.
This observation is consistent with \citet{zhou-waibel-2021-family}.
We attribute this to \lang{af}'s relationship with \lang{en}: both are Germanic and share the Latin script, with large lexical overlap. 
Multilingual machine translation systems can learn shared representations for linguistically similar languages \citep{dabre2017empirical,neubig-hu-2018-rapid, kudugunta-etal-2019-investigating,hokamp2019evaluating}; we expect that mBART taps into this relationship. Further, a smaller token set may help explain this improved generalization \citep{arivazhagan2019massively}.

For unseen languages that share the Latin script with English, explaining mBART's performance is less trivial, so we turn to a computational analysis. \lang{ga} reaches lower BLEU than \lang{yo}, despite being Indo-European like most of mBART's training data. It could be a result of its rare VSO word order \citep{Liu_Winata_Cahyawijaya_Madotto_Lin_Fung_2021}, its initial consonant mutations, or other rare syntactic phenomena. To explain the divergent behavior of \lang{af} and \lang{ga}, we use syntactic features estimated by the \(k\) nearest neighbors \citep{littell-etal-2017-uriel} of their WALS features \citep{wals}. \Cref{fig:syn_sim} shows the syntactic similarities between \lang{af}, \lang{ga}, and four high-resource languages (\lang{en}, \lang{de}, \lang{fr}, and \lang{nl}). This confirms that \lang{af} is more syntactically similar to these high-resource languages than \lang{ga} is.

Finally, we consider the interplay of translation direction and BLEU. Translating into \lang{en} regularly outperforms translating from \lang{en}, which we may attribute to mBART and the Transformer learning a strong \lang{en} language model in the decoder \citep{voita-etal-2021-analyzing}. But it may also come from BLEU's ignorance of subword phenomena. When translating into a morphologically rich language like \lang{si} or \lang{ta},  no partial credit is awarded for partially correct sets of morphemes. We see this as bolstering the movement toward character-aware metrics \citep{popovic-2015-chrf,mager-etal-2021-findings}.

\section{Conclusion}
We have assessed the value of \coolabbrev{} models like mBART for low-resource machine translation. 
We designed a reusable framework of experiments, capturing mBART's sensitivity to five facets of data.
Consistently, mBART fails in learning to translate new under-resourced languages---those unseen in the pre-trained model.
For languages used in monolingual pre-training, we find four- to tenfold data efficiency over a from-scratch Transformer, plus robustness to domain differences. 

For domain-specific datasets, mBART might outperform standard Transformers by an efficiency of five to ten times; 
future work can pinpoint the saturation size.
Fine-tuned mBART is robust to domain differences, while the Transformer flounders for out-domain datasets. 
However, the performance on unseen languages is generally not indicative of usable translation system.

Taken in tandem, these results point to the paramountcy of monolingual pre-training for the bilingual task of translation.
The biggest open issue, though, is not how to tune \coolabbrev{} models on limited data; instead, \textit{greater data acquisition} is the hope for truly low-resource machine translation.



\section*{Acknowledgments}
This project has been supported by the ICLR  CoSubmitting Summer (CSS) program 2022 initiated by ICLR DEI co-chairs Rosanne Liu and Krystal Maughan. David Adelani acknowledges the support of the EU funded Horizon 2020 project ROXANNE under grant agreement No. 833635. Lastly, we thank the Spoken Language Systems Chair, Dietrich Klakow at Saarland University for providing GPU resources to train the models.
\flushcolsend  
\clearpage

\bibliography{anthology,custom}
\bibliographystyle{acl_natbib}

\clearpage
\appendix
\section{Supplementary Material on Corpora} \label{sec:data}

Here we give details of the corpora used in our study.

\paragraph{Bible.} The JHU Bible Corpus \citep{mccarthy-etal-2020-johns} is a recently released corpus of Bible translations in over 1600 languages. In several low-resource languages, the Bible is the only available text parallel with another language; moreover, its verse structure makes it multi-parallel across thousands of languages. It has been used to assess multilingual translation at massive linguistic scale \citep{mueller-etal-2020-analysis}, develop new morphological tools \citep{nicolai-etal-2020-fine}, and fine-tune pre-trained language models to new low-resource languages \citep{ebrahimi-kann-2021-adapt}.

\paragraph{Gov't.} The government document corpus of \citet{fernando2020data} is a multilingual corpus for Sinhala, Tamil, and English. It contains official Sri Lankan government documents: annual reports, crawled content from government institutional websites, committee reports, procurement documents, and acts.

\paragraph{PMI.} PMIndia \citep{Barry2020PMIndia} is a parallel corpus of news updates for English and 13 other languages in India, extracted from the Prime Minister of India's website.

\paragraph{JW300.} The JW300 corpus \citep{agic-vulic-2019-jw300} is another parallel corpus, spanning 343 languages. It is obtained from \url{jw.org} and includes Jehovah's Witness magazines like \textit{Awake} and \textit{Watchtower}. The domain is highly religious, but it includes other societal topics such as reports about persecution of their disciples around the world. While JW300 was automatically aligned, \citet{abbott2019benchmarking} and \citet{alabi-etal-2020-massive} have verified its quality for African languages. For languages with non-Latin scripts in our study, the alignment has been judged to be poor by native speakers.

\paragraph{DGT.} The European Commission's Directorate-General for Translation--Translation Memory \citep{tiedemann2012parallel} covers 25 languages and corresponds to the `Summaries of EU legislation'. They are short explanations of the main acts passed by the European Union. The legislation included in the dataset includes directives, regulations, decisions, and international agreements.

\paragraph{Common Crawl.} CCAligned \citep{el2020massive} and CCMatrix \citep{schwenk-etal-2021-ccmatrix} are web-scraped corpora that were automatically aligned using LASER sentence embeddings \citep{schwenk-2018-filtering}. CCAligned is newer, and it has more text in low-resource languages. The dataset, albeit noisy \citep{qualityAtAGlance}, has been used to develop highly multilingual machine translation models like M2M100~\cite{fan2021beyond} and mBART multilingual MT~\cite{tang2020multilingual}; a modified version is used to train mT5 \citep{xue2021mt5}.

\paragraph{Data splits}
For \flores{} and the Bible, we always use 1000 sentence pairs for development \citep[see][]{kann-etal-2019-towards} and 1000 sentence pairs for test. For the second in-domain dataset, the size varies between 1000 and 2000 sentence pairs based on availability.

\section{Supplementary Material on Experimental Setup} \label{app:experiments}

\paragraph{mBART and mT5.}
We compared mBART50 and \texttt{mT5-base} because they have comparable numbers of parameters.
For both the mBART50 and \texttt{mT5-base} models \citep{tang2020multilingual}, we train up to 3 epochs with a learning rate of \(5 \times 10^{-5}\), dropout of 0.1, maximum lengths of 200 for the source and target, and a batch size of 10. We decode using beam searh with a beam size of 5. We use the implementations in the HuggingFace Transformers library, and we leverage hardware-level parallelism by training on \textsc{Nvidia} Tesla V100 GPUs.

We perform bilingual fine-tuning on the 10 selected language pairs. For each language direction, we initialize the encoder--decoder model's parameters from the pre-trained mBART model's corresponding encoder and decoder. After initialization, we continue training. 

Because mBART requires a target language to be specified during decoding from amongst those that the model has seen, we follow past work in selecting languages related to our target languages for unseen languages \citep{madaan-etal-2020-transfer,cahyawijaya2021indonlg}. Considering syntactic and phylogenic closeness of languages \citep{wals,littell-etal-2017-uriel}, we chose \lang{bn} for \lang{as}, \lang{te} for \lang{kn}, \lang{fr} for \lang{ga}, and \lang{sw} for \lang{yo}.

\paragraph{mT5.} Considering memory bottlenecks, we use the \texttt{mT5-base} model. It supports over 100 languages, including five of the six from our preliminary experiment. Because Irish (\lang{ga}) is not among these, we use the French language code for fine-tuning the model.

\paragraph{Transformer.} We train Transformer models implemented in \textsc{Fairseq} using the same datasets as we used for fine-tuning mBART. We use two Transformer architectures, depending on the data size. When there are fewer than 10k parallel sentences, the model consists of 3 encoder layers and 3 decoder layers, with embedding dimension of 512 and 2 attention heads. When there are 10k or more parallel sentences, we instead use a model that consists of 6 encoder layers and 6 decoder layers, with an embedding dimension of 256 and 2 attention heads. In each case, we have an initial learning rate of \(1\times 10^{-3}\), a weight decay of \(1 \times 10^{-4}\), dropout of \(0.4\), and batch size of \(32\). We use early stopping based on the validation loss.
We train the models from scratch with segmentation into subword tokens performed by SentencePiece. When decoding, we use beam search with a beam size of \(5\).

\paragraph{Evaluation.}
To ease the comparison of future work with ours, we report that the SacreBLEU settings we use are represented by the signature \verb|BLEU+c.mixed+#.1+s.exp+tok.13a+v.1.5.0|.

\clearpage

\end{document}